\documentclass[10pt,twocolumn,letterpaper]{article}

\usepackage{iccv}
\usepackage{times}
\usepackage{epsfig}
\usepackage{graphicx}
\usepackage{amsmath}
\usepackage{amssymb}

\usepackage{makecell}
\usepackage{multirow,xcolor}
\usepackage{subcaption}

\definecolor{citecolor}{RGB}{0,0,0}
\usepackage[pagebackref=true,breaklinks=true,letterpaper=true,colorlinks,
citecolor=citecolor,bookmarks=false]{hyperref}

\iccvfinalcopy 

\def\iccvPaperID{****} 

\ificcvfinal\fi
\begin{document}

\title{FAB: A Robust Facial Landmark Detection Framework\\
for Motion-Blurred Videos}

\author{Keqiang Sun$^{1}$, Wenyan (Wayne) Wu$^{2}$, Tinghao Liu$^{3}$, Shuo Yang$^{4}$\\
Quan Wang$^{3}$, Qiang Zhou$^{2}$, Zuochang Ye$^{1}$, Chen Qian$^{3}$\\
$^{1}$Institute of Microelectronics, Tsinghua University\\
$^{2}$Beijing National Research Center for Information Science and Technology (BNRist), Tsinghua University\\
$^{3}$SenseTime Research  
$^{4}$Amazon Rekognition\\
\tt\small {\{skq17,wwy15\}@mails.tsinghua.edu.cn,
\{zuochang,zhouqiang\}@tsinghua.edu.cn,}\\
\tt\small{\{liutinghao,wangquan,qianchen\}@sensetime.com,
shuoy@amazon.com}
}

\maketitle
\thispagestyle{empty}

\begin{abstract}
Recently, facial landmark detection algorithms have achieved remarkable performance on static images. However, these algorithms are neither accurate nor stable in motion-blurred videos. The missing of structure information makes it difficult for state-of-the-art facial landmark detection algorithms to yield good results.

In this paper, we propose a framework named FAB that takes advantage of structure consistency in the temporal dimension for facial landmark detection in motion-blurred videos. A structure predictor is proposed to predict the missing face structural information temporally, which serves as a geometry prior. This allows our framework to work as a virtuous circle. On one hand, the geometry prior helps our structure-aware deblurring network generates high quality deblurred images which lead to better landmark detection results. On the other hand, better landmark detection results help structure predictor generate better geometry prior for the next frame. Moreover, it is a flexible video-based framework that can incorporate any static image-based methods to provide a performance boost on video datasets. Extensive experiments on Blurred-300VW, the proposed Real-world Motion Blur (RWMB) datasets and 300VW demonstrate the superior performance to the state-of-the-art methods. Datasets and models will be publicly available at \href{https://keqiangsun.github.io/projects/FAB/FAB.html}{https://keqiangsun.github.io/projects/FAB/FAB.html}.
\end{abstract}

\vspace{-0.1cm}

\section{Introduction}

\begin{figure}[t]
\centering
\includegraphics[width=1.0\linewidth]{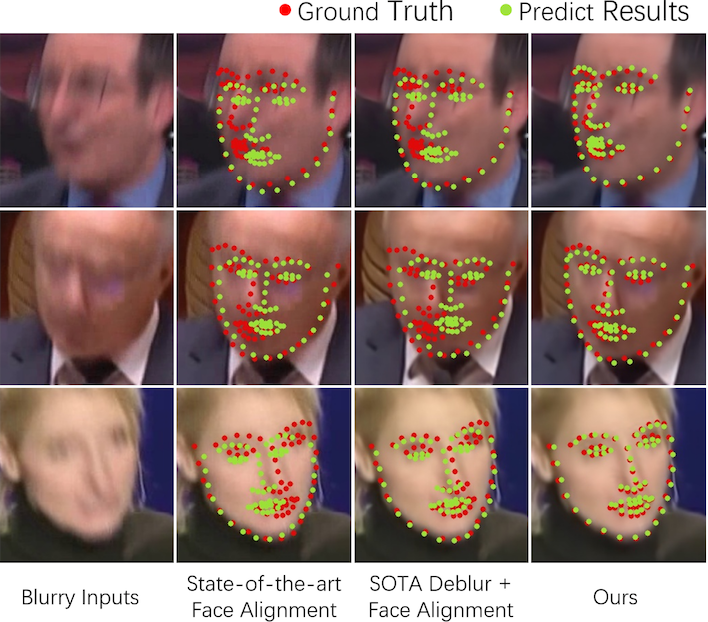}
\caption{\small{The first column is the frames of Blurred-300VW. In the second column, the results of state-of-the-art facial landmark detection algorithm are given. The third column corresponds to a naive combination of deblurring and facial landmark detection. In the fourth column are the results of our proposed algorithm.}}
\label{fig:cmp4}
\vspace{-0.4cm}
\end{figure}

Facial landmark detection, or known as face alignment, serves as a key component for many face applications, \eg face recognition, face verification and face augmented reality. Previous researches~\cite{xiong2013supervised,zhang2014facial,zhu2015face,DVLN,dong2018style,dong2018supervision,wu2018look,merget2018robust} mainly focus on detecting facial landmarks in static images. One natural extension of image-based facial landmark detection is video facial landmark detection, which tries to locate facial landmarks in sequential frames. Different from static images, motion blur usually appears in videos, due to the mismatch of the motion speed and shutter closing speed. As shown in Figure~\ref{fig:cmp4}, the missing of structure information, \ie edges or boundaries, in the motion-blurred videos makes it difficult for the state-of-the-art facial landmark detection algorithms to capture facial structure. The objective of this paper is to devise an effective framework to handle facial landmark detection in motion-blurred videos.

An intuitive method is to employ a deblurring algorithm before facial landmark detection. It is straightforward that the deblurred image would promote facial landmark detection performance. State-of-the-art face deblurring algorithms~\cite{shen2018deep,chrysos2017deep,huang2015efficient} tend to rely on facial structure (\eg facial landmarks and edges) from the input image as strong priors to restore the shape and details. Nevertheless, these face structure would be unavailable when the landmarks cannot be precisely predicted,~\eg in extremely blurry frames.
In conclusion, the facial landmark detection in motion-blurred videos requires deblurred images, while face deblurring tends to rely on face structure like facial landmarks. They are mutually beneficial and interdependent, which makes it a ``chicken or egg'' dilemma in motion-blurred videos.

In a video clip, face structure keeps temporal continuity and consistency, and blurry frames usually intersperse along the time dimension. This motivates us to predict a reliable face structure according to previous structural information. Inspired by this idea, we proposed a Structure Predictor to predict current face structure. Specifically, given previous facial edges, the predictor figures out the optical flow and extends the motion to predict next face edges by assuming the optical flow is linear.

\begin{figure*}[t]
\centering
\includegraphics[width=0.9\linewidth]{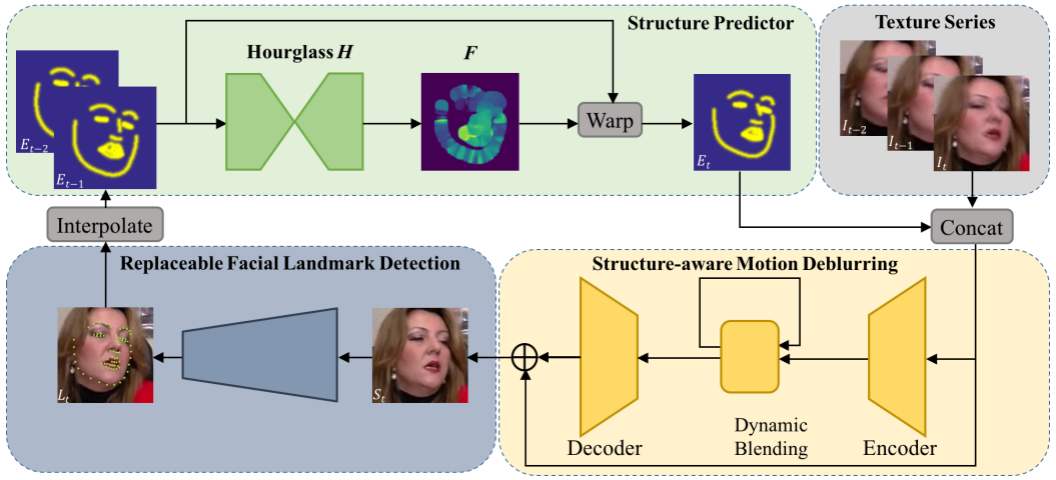}
\caption{\small{Framework. Given two previous face edges $E_{t-2}$ and $E_{t-1}$, hourglass $\pmb H$ predicts the optical flow $F$ between the two boundary maps. Warping block $\pmb W$ warps $E_{t-2}$ and $E_{t-1}$ into next boundary map $E_t$ according to the predicted optical flow $F$.
Recent frames $I_{t-2}, I_{t-1}, I_t$, concatenated with the predicted boundary map, are feed to the Boundary-aware Deblur Network $\pmb D$, which produces a sharp face $S_t$.
Taking the deblurred sharp image $S_t$ as input, the replaceable facial landmark detection network predicts more accurate landmark location $L_t$, which is interpolated to face edges and provides the face structural information for next loop.}}
\label{fig:framework}
\vspace{-0.4cm}
\end{figure*}

Based on this, we designed a framework, composed of three modules, \ie structure predictor, structure-aware motion deblurring network, and replaceable facial landmark detection network, as is depicted in Figure~\ref{fig:framework}. These three components work as an organic whole. The structure predictor predicts the current facial structure from previous temporal cues. The deblurring network removes the motion blur with the assistance of the structure prior. With the deblurred image, the landmark detector yields accurate landmarks, which are used to predict the next face structure.

Since our framework is proposed for Face Alignment in Blurred-videos, we call it ``FAB''. FAB is designed with the advantage of great flexibility. Each component of the framework can be replaced with faster or more accurate backbone, which allows our framework to be updated easily and meet more demands. 

For a better evaluation, we proposed Blurred-300VW and RWMB dataset with severe artificial and real-world motion blur respectively. Extensive experiments demonstrate the effectiveness of our framework and the superior performance to state-of-the-art methods on both datasets. In conclusion, the contributions of this paper are:

(1) A framework, in which a face deblurring network and landmarks detector work as a \textit{virtuous circle} and obtain state-of-the-art performance in motion-blurred videos.

(2) A novel component, structure predictor, which utilizes temporal information to provide reliable face structure.

(3) Two new datasets, (\ie, Blurred-300VW and RWMB) which are more suitable benchmarks for video facial landmark detection tasks.

\section{Related work}

\subsection{Facial Landmark Detection}
\noindent
\textbf{Facial landmark detection in static images.}
The classic model-based methods ASMs~\cite{milborrow2008locating}, AAMs~\cite{cootes2001active,kahraman2007active,matthews2004active,saragih2007nonlinear}, CLMs~\cite{kumar2008facetracer,saragih2011deformable}, ESR~\cite{cao2014face}, SDM~\cite{xiong2013supervised}, CFSS~\cite{zhu2015face}, and deep convolutional neural wetwork methods,~\eg TCDCN~\cite{zhang2014facial}, FAN~\cite{bulat2017far}, DSRN~\cite{miao2018direct}, RFLD~\cite{merget2018robust}, SAN~\cite{dong2018style}, LAB~\cite{wu2018look} have obtained increasingly excellent performance in static images under different poses, light conditions, expressions, etc. However, few works in the literature of face alignment pay attention to the motion blur. They cannot maintain outstanding performance in case of severe motion blur. Our paper looks into blurry scenario and propose a framework to remove the motion blur, and then promote all these state-of-the-art facial landmark detection algorithms.

\noindent
\textbf{Facial landmark detection in videos.}
To overcome challenging problems like large pose and occlusion, facial landmark detection is naturally extended to videos leveraging temporal information~\cite{chrysos2018comprehensive}. Xi \etal~\cite{peng2016recurrent} propose a recurrent encoder-decoder network (RED), combined with spatial and temporal recurrent learning, to explicitly model the temporal dependency relationship on frames. Hao \etal~\cite{liu2018two} construct the two-stream transformer network (TSTN), where the temporal stream learns to capture the continuous consistency across multiple frames in video clips and the spatial stream is capable of locating landmarks. 
These~\cite{burgos2013merging,shen2015first,peng2016recurrent,gu2017dynamic,liu2018two,chrysos2018pd2t} are typical works employing temporal information in facial landmark detection. However, few works have ever noticed the motion blur, a common problem and challenge in most videos. Moreover, face structure information has not been paid attention to in these video-based methods. In our paper, we propose a framework that leverages the temporal and structural information to tackle the problem of motion blur. 

\subsection{Motion Deblurring}

Motion blur, which usually happens for the mismatching of the motion speed and shutter closing speed, could be entailed with object movement, camera shake, etc. The unlimited nature of motion makes deblurring a complex problem. Recent years have witnessed great progress in general image deblurring~\cite{xu2010two,xu2013unnatural,kupyn2017deblurgan,park2017joint,wieschollek2017learning,zhang2018dynamic,Chrysos2018MotionDO}. Compared with image deblurring, video deblurring~\cite{hyun2015generalized, kim2017online, su2017deep} can utilize temporal information to handle large motion blur with fewer network parameters. 
However, these methods are not specialized for face and therefore have not leveraged face structural information.

Structure information could efficiently assist deblurring.
Prior knowledge, especially facial structure~\cite{huang2015efficient,bulat2017super,shen2018deep}, has been proven to be an effective face prior in corresponding tasks such as super-resolution and deblurring.
However, as mentioned in~\cite{shen2018deep}, these methods fail when the input face images are not well aligned, \eg side faces or extremely large motion where semantic face parsing or landmark detection fails. 
Considering all these drawbacks, we designed the framework, which exploits temporal information to provide reliable structure information and furthermore removes the blur deeply.

\section{Method}
\vspace{-0.1cm}
As shown in the figure~\ref{fig:framework}, the proposed FAB can be divided into three conjoined components: structure predictor, structure-aware deblurring network and replaceable facial landmark detection network. The structure predictor predicts facial structure prior for the current frame from previous frames. With help from the predicted facial structure prior, the structure-aware deblurring network generates a clear image. Given a clear input image, the facial landmark detection network produces accurate facial landmarks for the current frame, which is fed back to the structure predictor for the next frame prediction. This makes the three networks an organic whole to perform facial landmarks localization and deblurring simultaneously and benefit each other in motion-blurred videos.

\subsection{\label{structure_predictor}Structure Predictor}
\vspace{-0.1cm}
Structure predictor plays a key role in breaking the "chicken or egg" dilemma between deblur and facial landmark detection problems. It takes input facial structure of time $t-2$ and $t-1$ to predicts the facial structure of current frame $t$. The motion caused by camera shake or object motion can be modeled mathematically and are continuous in a short time. Moreover, facial structure is a semantic meaningful, clear and well-defined representation of face regardless of the face texture. The motion between two facial structures is much easier to extract than motion between two face images. These properties make short time facial structure prediction a feasible problem.

A facial structure could be represented by landmarks, edges, part segments, 3D models \etc. Landmarks are not semantically stable as they would drift along edges under different expressions and poses. The drift of landmarks adds noises to face motion which is harmful in structure prediction. On the other hand, complex annotation such as part segments and 3D models contains richer information but the large-scale dataset is hard to obtain. Thus we use face edges as facial structure information. The edges are interpolated from landmarks of each facial component.

Given two previous face edges $E_{t-2}$ and $E_{t-1}$, following previous paper~\cite{liu2017video}, we use an hourglass $\pmb H$ to predict the optical flow $F$ between the two boundary maps. Directly predicting face edges $E_t$ is also feasible but can hardly guarantee the sharpness property of the boundary map. However, this property is essential for following the deblur network. Then the warping block $\pmb W$ would warp $E_{t-2}$ and $E_{t-1}$ into the next boundary map $E_t$ according to the predicted optical flow $F$:
\vspace{-0.1cm}
\begin{equation}
\begin{aligned}
\pmb H(E_{t-2},E_{t-1}) = \pmb F 
\end{aligned}
\label{equation_1}	
\end{equation}
\begin{equation}
\begin{aligned}
\pmb W(E_{t-2},E_{t-1},\pmb F) = E_t
\end{aligned}
\label{equation_2}
\end{equation}
where $t$ means the current time.

The facial structure predictor is pre-trained using facial landmark detection video dataset like 300VW, then fine-tuned together with other networks. For pretraining, the mean squared error (MSE) between the predicted and ground truth face boundary map. The loss is defined as

\begin{equation}
\begin{aligned}
\mathcal{L}_{prd} = \frac{1}{N_{pixel}}\cdot{\left \| E_t - E_{GT} \right \|_2}
\end{aligned}
\label{equation_3}
\end{equation}
Where $N_{pixel}$ means the total pixel number in the generated image, and $E_{GT}$ is the ground truth of the current face edges. In the final experiment, we show that the motion of edges can be predicted accurately.

\subsection{Structure-Aware Motion Deblurring}
\begin{figure}[t]
\centering
\includegraphics[width=0.9\linewidth]{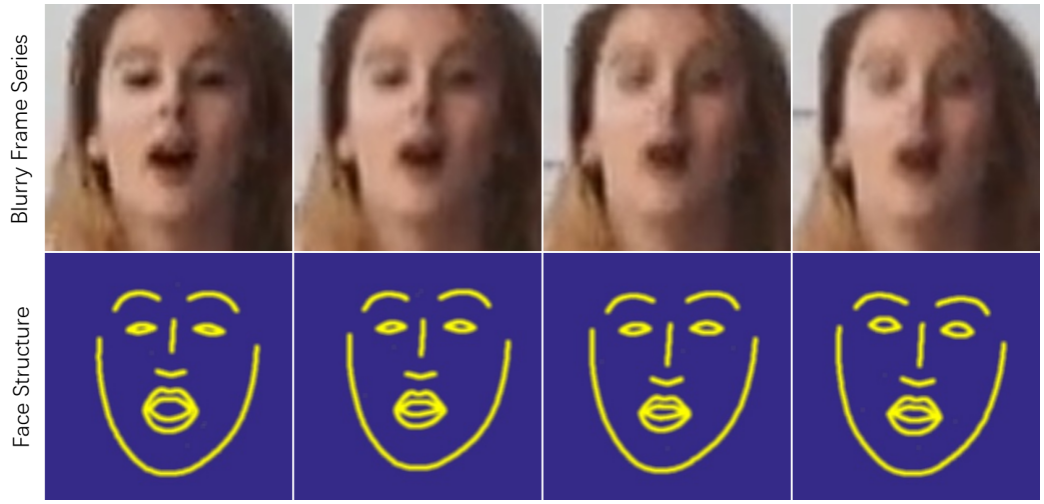}
\caption{\small{Face structural information is reliable. In the first row, the face remains relatively static to the bounding box, but gradually getting blurry. However, as is shown in the second row, face edges are exempt from these problems, and thus much easier to predict.}}
\label{fig:boundary_reliable}
\end{figure}

Face deblurring is difficult itself. Face motion is inconsistent with the motion blur degree because of the use of face detection/tracking. As is shown in figure \ref{fig:boundary_reliable}, the face remains relatively static to the bounding box, but gradually getting blurry. However, face edges are exempt from these problems(see the second row of figure \ref{fig:boundary_reliable}), and thus much easier to predict. With the reliably predicted structure prior, the deblurring network entangles the textural and structural information and reconstructed a deblurred face.

We disentangled a face into two parts, \ie structure, and texture. Given accurate structure prior, boundary map, the deblur network $\pmb D$ reconstructs a sharp image, which meets the boundary constraint and maintains the texture consistency simultaneously. Face structural information guides the motion deblurring of current frame. This process could also be viewed as filling the predicted structure with the texture from previous frames.
Following state-of-the-art deblur network design~\cite{kim2017online,wieschollek2017learning}, an batch of three recent frames, $I_{t-2},I_{t-1},I_t$ are used as input. Then we concatenate the predicted boundary map, as structure prior, with the batch of frames as input to our deblur network. An encoder network extracts needed information from inputs. Then we use a dynamic temporal blending network, following~\cite{wieschollek2017learning} and~\cite{kim2017online}, to combine information across different frames. Finally, a decoder is used to predicts the residual between blurry frames and groundtruth.
\begin{equation}
\begin{aligned}
\pmb D(E_t,I_{t-2},I_{t-1},I_t) = S_t
\end{aligned}
\label{equation_4}
\end{equation}

Similar to facial structure predictor, we also pre-train structure-aware deblur network using 300VW. First, blurred videos are generated by using methods from ~\cite{kim2017online}. Then the ground-truth edges are used as structure prior to pre-train this network. We employ the mean squared error $\mathcal{L}_{rec}$ between the deblurred image and the ground truth sharp image:

\begin{equation}
\begin{aligned}
\mathcal{L}_{rec} = \frac{1}{N_{pixel}}\cdot{\left \| S_t - S_{GT} \right \|_2}
\end{aligned}
\label{equation_5}
\end{equation}

\subsection{Replaceable Facial Landmark Detection Network}
By using the deblurred sharp image $S_t$ as input, the replaceable facial landmark detection network predicts the landmark location $L_t$. The facial landmark detection network can be any advanced network architecture. In this paper, we use the residual network with pre-activation~\cite{he2016identity,he2016deep}. And L1 distance $\mathcal{L}_{align}$ is used as the loss for facial landmark detection. The network is first pre-trained on facial landmark detection dataset and fine-tuned together with the other two networks:
\begin{equation}
\begin{aligned}
\mathcal{L}_{align} = \frac{1}{N_{point}}\cdot{\left \| L_{t} - L_{GT} \right \|_1}
\end{aligned}
\label{equation_6}
\end{equation}
where $L_{GT}$ means the ground truth facial landmark location. $N_{point}$ is the number of landmarks.

The deblurred frames would lead to better performance for facial landmark detection task regardless of what the facial landmark detection network is used in our framework. Almost every single landmark localization networks would have much better performance in motion-blurred videos. Moreover, the more accurate facial structure would further provide better information for structure predictor in the next loop, which forms a virtuous circle.

\begin{figure*}[t!]
\centering
\includegraphics[width=1.0\linewidth]{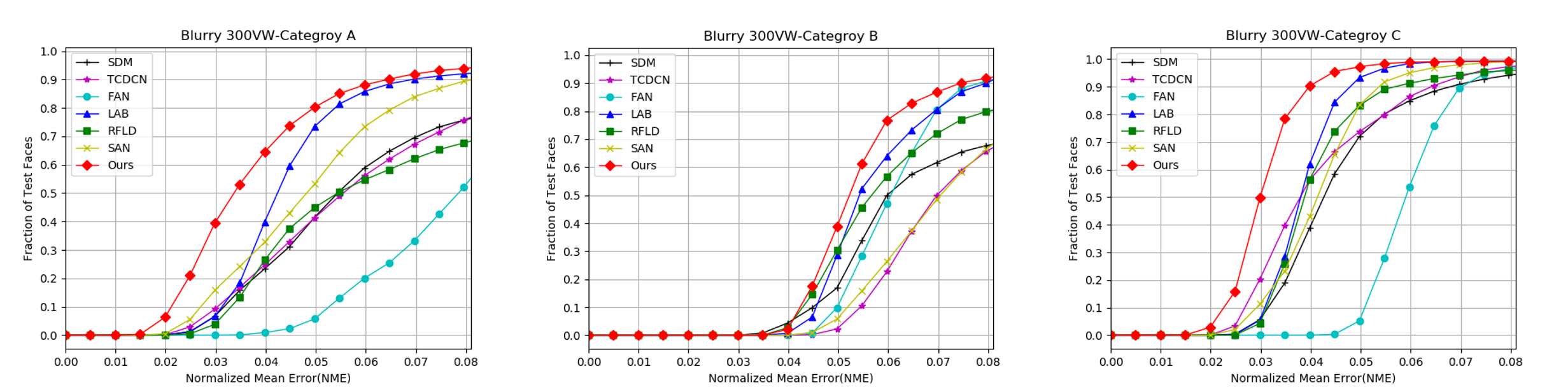}
\caption{\small{CED for Blurred-300VW Testset.}}
\label{fig:ced_blur}
\vspace{-0.3cm}
\end{figure*}

\subsection{Alternate Fine-Tuning}
Structure predictor, structure-aware motion deblurring network and facial landmark detection network mentioned above constitute our main architecture. 
The three networks are pretrained separately and finally alternately end-to-end fine-tuned together to minimize the total loss:
\begin{equation}
\begin{aligned}
\mathcal{L}_{total} = \mathcal{L}_{str} + \mathcal{L}_{rec} + \mathcal{L}_{align}
\end{aligned}
\label{equation_7}
\end{equation}
where the $\mathcal{L}_{str}$ $\mathcal{L}_{rec}$ $\mathcal{L}_{align}$ are calculated according to Equation~\ref{equation_3},~\ref{equation_5} and~\ref{equation_6} respectively.

During the end-to-end fine-tuning, the three networks benefit each other to achieve better performance. The losses of other networks are also helpful. For example, with better deblur results, the alignment network can achieve higher accuracy. Thus the loss of facial landmark detector also helps optimizing face deblur network. The gradient of facial landmark detector would propagate back to encourage sharper input as well as pre-stage deblur network.

\vspace{-0.1cm}
\section{Implement Details}

\subsection{Network Structure}
Our proposed algorithm is composed of three blocks, namely structure predictor, structure-aware motion deblurring network, and a replaceable facial landmark detection network. 
We used eight residual blocks in total to build the Hourglass~\cite{DBLP:journals/corr/NewellYD16} in the Structure Predictor. 
There are two convolutional layers and four residual blocks correspondingly in the Encoder and Decoder in the structure-aware motion deblurring network. 
Since the purpose of our work lies on the overall system framework instead of the design of the network or the loss function, we employed a simple network (\ie pre-activated Resnet-18~\cite{he2016identity}) as our replaceable facial landmark detection network. We present detailed information about the network structure in the first section of the supplementary material.

\subsection{Training}
All the three components are required to be pretrained respectively. For structure predictor, facial edges are implemented as inputs.
We fit annotated landmarks face edges by cubic spline interpolation~\cite{mckinley1998cubic} and use them as input and ground truth.
For structure-aware motion deblurring network, annotation results for the current frame are interpolated to face edge and used for structural information.
For facial landmark detection network, we trained the model on 300W dataset~\cite{sagonas2013300}, then finetune the network on the 300VW dataset~\cite{shen2015first}. Data augmentation such as translation, rotation, flipping, and zooming is also used in the total training stage.

During the end-to-end training, we designed an alternate training method to obtain better reconstruction and landmark localization results.
Facial landmark detection network and the structure predictor take turns to be trained for one epoch. In this way, the landmark detection network would take deblurred images as input instead of original images.

\vspace{-0.1cm}
\section{Experiments}

\subsection{Datasets and Evaluation Metric}

\noindent
{\bf 300W~\cite{sagonas2013300} and 300VW~\cite{shen2015first}.}
300W and 300VW are popular benchmarks for facial landmark localization methods. 300W contains 3,148 training images and 689 testing images. 300VW is designed as a benchmark for videos, containing 50 training videos and 64 testing videos. The testing set of 300VW is divided into three scenarios, \ie category A: well-lit arbitrary expressions, category B: unconstrained illuminations and category C: arbitrary conditions. In this paper, these two datasets are combined in training.

\begin{figure}
\begin{minipage}{0.42\linewidth}
 \centerline{\includegraphics[width=1.3\linewidth]{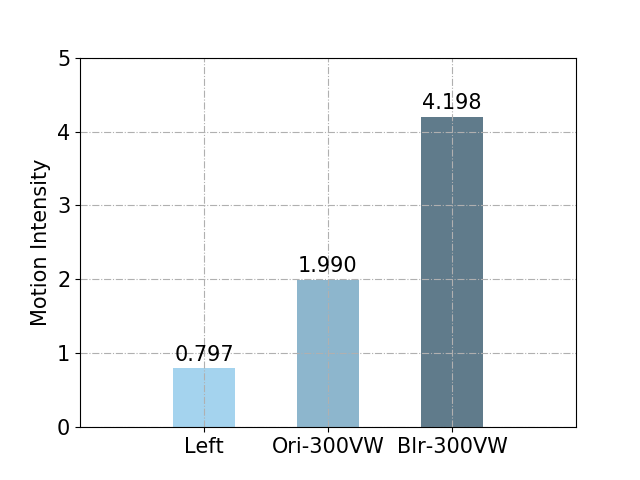}}
 \centerline{(a) Motion Intensity.}
\end{minipage}
\hfill
\begin{minipage}{0.42\linewidth}
 \centerline{\includegraphics[width=1.3\linewidth]{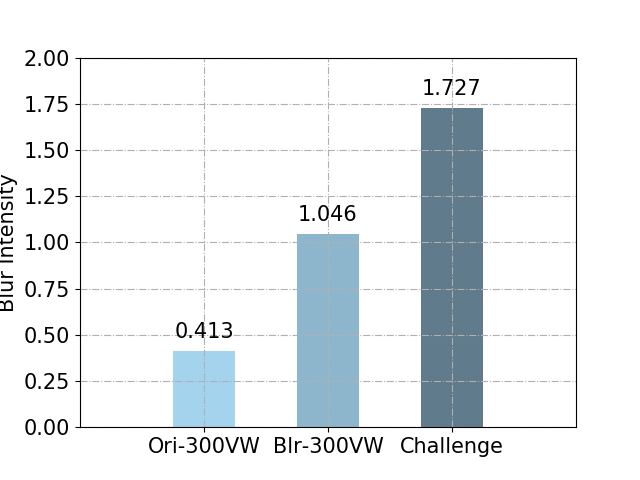}}
 \centerline{(b) Blurry Intensity.}
\end{minipage}
\caption{\small{(a) Comparison of the motion intensity among the left videos of original 300VW, images of original 300VW and images of blurred-300VW. (b) Comparison of the blur intensity among the original 300VW, Blurred-300VW and Challenge Subset.}}
\label{fig:motion-blur-intensity}
\end{figure}

\noindent
{\bf Blurred-300VW.}
In order to generate more severe motion blur, larger motion in original videos is a need. We select subsets (31 videos in the training set and 9 videos in the testing set) from 300VW with large motion according to a face motion intensity index.
The face motion intensity is defined by accumulating the movement of the left eye during a time unit and normalizing it with the inter-ocular distance. As is shown in Figure~\ref{fig:motion-blur-intensity} (a), the selected videos are characterized with far more severe motion intensity. 

Then, we blurred these picked subsets, following the method in~\cite{kim2017online}. For each adjacent three frames, we interpolated them with 20 subframes according to the optical flow~\cite{horn1981determining}. The mean value of these 20 subframes is calculated to mimic motion blur. Annotation of each generated frame is taken from the middle-time subframe. The blurred-300VW dataset contains obvious motion blur especially when the face moves greatly, which is suitable for the illustration of this work. 

\begin{figure}[t]
\centering
\includegraphics[width=0.9\linewidth]{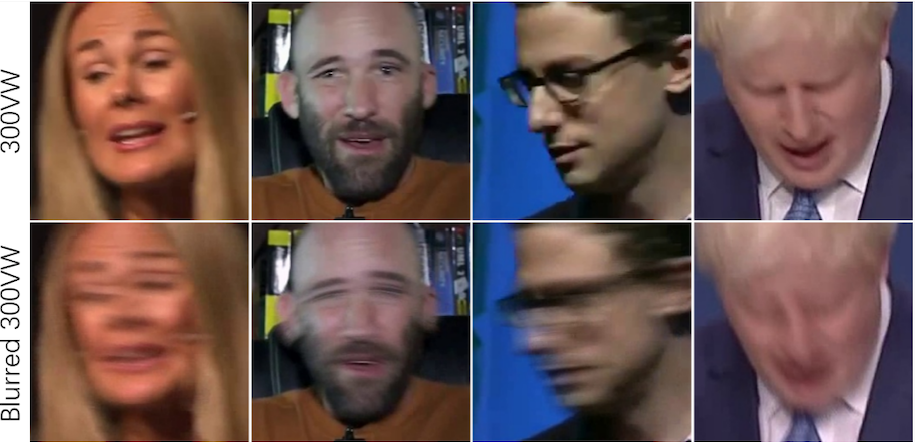}
\caption{\small{Comparison of random sampled corresponding images on original 300VW and Blurred-300VW.}}
\label{fig:sharp-blur}
\vspace{-0.4cm}
\end{figure}

Moreover, to evaluate the limit of recent facial landmark detection methods, we further form a Challenge Subset, which contains hundreds of images from 6 test videos with extreme motion blur. The comparisons of the blurry degree for these three sets are shown in Figure~\ref{fig:motion-blur-intensity} (b). Also, we compare images on 300VW and Blurred-300VW at the same time in Figure~\ref{fig:sharp-blur}.

\noindent
{\bf RWMB.}
Real-World Motion Blur (RWMB) is our newly proposed in-the-wild benchmark for facial landmark detection task in realistic motion-blurred videos. It currently contains 20 videos with obvious real-world motion blur picked from YouTube, which include dancing, boxing, jumping, \etc. 
There are $35,540$ frames, which are all annotated with 98 landmarks following the protocol of WFLW~\cite{wu2018look}. Moreover, RWMB dataset will be further enlarged to hundreds of videos, including millions of frames. Please notice the news in the project page for more information.

Even though human beings could generally recognize the position of facial features of blurry faces, it is challenging to determine the specific location of each landmark. The annotation of the previous frame is presented to the annotator as a reference. Each frame is annotated by three expert annotators and checked by two quality inspector. As motion blur exists widely in regular videos, research in this scenario is challenging and meaningful. This benchmark aims at promoting the development of facial landmark detection in motion-blurred videos.

\noindent
\textbf{Annotation protocol.}
We provided two versions of annotation of RWMB and 300VW, \ie 98 landmarks following~\cite{wu2018look} and 68 landmarks following~\cite{shen2015first}. In this way, we unify these two video facial landmark datasets and hope to facilitate the cross-dataset evaluation in future works.

\noindent
\textbf{Evaluation metric.}
To evaluate the performance of facial landmark detection, Normalized Mean Error (NME), Cumulative Errors Distribution (CED) curve, Area Under the Curve (AUC) and Failure Rate are employed in this paper. Mean Error is normalized by inter-ocular distance (namely outer eye corner distance) throughout this paper.

\subsection{Results}

\subsubsection{Evaluation on Blurred-300VW}
\label{sec:blurred-300vw}

\begin{table}
\begin{center}
\resizebox{0.95\columnwidth}{!}{
\begin{tabular}{c|c|c|c|c}
\Xhline{1.2pt}
Method & CateA & CateB & CateC & Year \\
\Xhline{1.2pt}
SDM~\cite{xiong2013supervised} & 10.4 & 8.52 & 4.83 & 2013 \\
TCDCN~\cite{zhang2014facial} & 6.81 & 7.90 & 4.36 &2016 \\
FAN~\cite{bulat2017far} & 10.84 & 6.80 & 6.11 &2017 \\ 
RFLD~\cite{merget2018robust} & 10.95 & 8.19 & 4.63 &2018 \\ 
SAN~\cite{dong2018style} & 5.62 & 7.71 & 4.35 &2018 \\ 
LAB~\cite{wu2018look} & 5.28 & 6.07 & 3.96 &2018 \\ 
\Xhline{1.2pt}
\textbf{Ours} & $\pmb {4.24} $ & $\pmb {5.67}$ & $\pmb {3.16}$ & - \\
\Xhline{1.2pt}
\end{tabular}
}
\end{center}
\caption{\small{NME(\%) value of our method and several state-of-the-arts on Testset of Blurred-300VW.}}
\label{nme_blurred_300VW}
\vspace{-0.3cm}
\end{table}

\begin{table*}[ht]
\begin{center}
\resizebox{2\columnwidth}{!}{
\begin{tabular}{c|c|c|c|c|c|c|c}
\Xhline{1.2pt}
Method & NME & \multicolumn{2}{c}{NME threshold: 0.2} & \multicolumn{2}{|c}{NME threshold: 0.1} & \multicolumn{2}{|c}{NME threshold: 0.08} \\
\cline{3-8}
~ & ~ & Failure Rate(\%) & AUC & Failure Rate(\%) & AUC & Failure Rate(\%) & AUC \\
\Xhline{1.2pt}
RFLD~\cite{merget2018robust} & 16.34 &25.67	&0.59	&44.84	&0.38	&54.39	&0.29	\\ 
FAN~\cite{bulat2017far} & 13.73 &10.42	&0.56	&34.18	&0.27	& 53.26	&0.18	\\ 
TCDCN~\cite{zhang2014facial} &10.48 &11.27	&0.60	&34.82	&0.41	&45.79	&0.34	\\
SAN~\cite{dong2018style} & 10.40 &11.79	&0.62	&31.37	&0.43	&40.93	&0.36	\\ 
LAB~\cite{wu2018look} & 9.47 &8.28	&0.61	&32.20	&0.43	&41.12	&0.34	\\
\Xhline{1.2pt}
\textbf{Ours} & \textbf{8.43}	&\textbf{5.25}	&\textbf{0.63}	&\textbf{28.77}	&\textbf{0.45}	&\textbf{38.54}	& \textbf{0.39}	\\ 
\Xhline{1.2pt}
\end{tabular}
}
\end{center}
\vspace{-0.3cm}
\caption{\label{rwmb_fr_auc_ocular}\small{Failure rate and Area under curve(AUC), normalized by inter-ocular distance, on RWMB dataset}}
\vspace{-0.2cm}
\end{table*}

\begin{figure*}[t]
\centering
\includegraphics[width=1.0\linewidth]{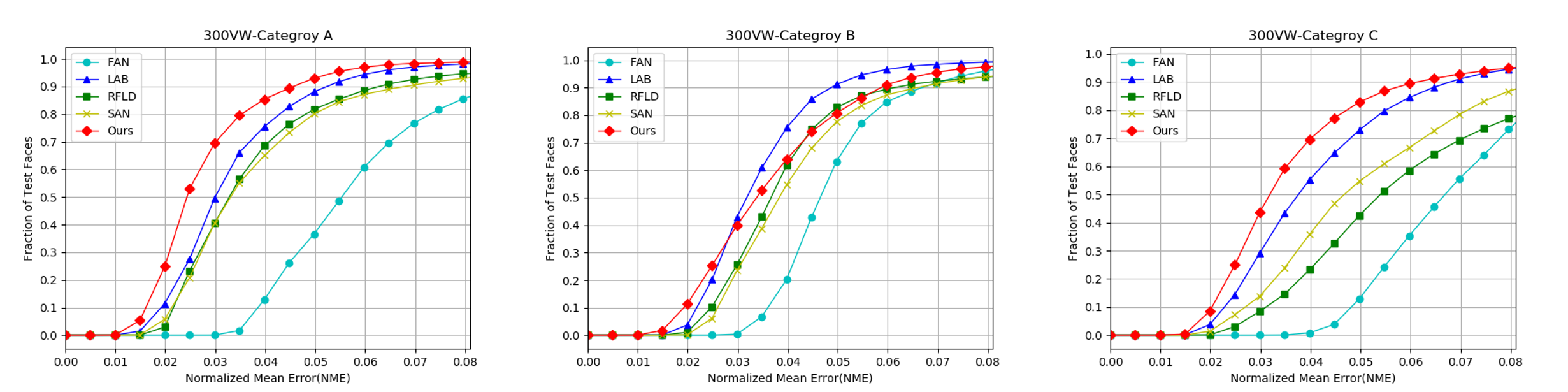}
\vspace{-0.3cm}
\caption{\small{CED for original 300VW Testset.}}
\vspace{-0.4cm}
\label{fig:ced_sharp}
\end{figure*}

This experiment is conducted to demonstrate that our algorithm is capable to handle \textit{artificial} motion blur, and accurately detect facial landmarks.

We re-implemented several typical state-of-the-art algorithms for comparison, including RFLD~\cite{merget2018robust}, SAN~\cite{dong2018style} and LAB~\cite{wu2018look}. For a fair comparison, we made great efforts to reproduce the comparable results reported in their papers.
The results on Blurred-300VW are shown in Table~\ref{nme_blurred_300VW}. The Experiment shows that our algorithm significantly outperforms previous methods by a large margin, and has improved the state-of-the-art performance (NME) by relatively 18.1\% on average on the blurred-300VW dataset from 5.28 to 4.24. Figure~\ref{fig:ced_blur} shows the CED curve of all these excellent algorithms. Note that SAN~\cite{dong2018style} and LAB~\cite{wu2018look} are rather robust in image-based facial landmark detection. However, when confronting the severe motion blur, they fail to yield reliable and robust facial landmarks due to the destruction of the structure information. Our method benefits from the reliable predicted facial structure and the deblurred input and obtains state-of-the-art performance.

\vspace{-0.2cm}
\subsubsection{Evaluation on RWMB}
\vspace{-0.2cm}

Although the experiments in Blurred-300VW have demonstrated that the proposed method could handle \textit{artifical} motion blur, we still wonder how our method performs when facing \textit{real-world} motion blur. Therefore, we propose this dataset, Real-World Motion Blur, as well as the testing result of several state-of-the-art algorithms to provide a benchmark.

We retrained several state-of-the-art algorithms as is introduced in Sec.~\ref{sec:blurred-300vw} and tested them on the 68-points version of the RWMB. Comparison results on the RWMB against these methods are presented in Table~\ref{rwmb_fr_auc_ocular}. The experiment shows that our algorithm has improved the state-of-the-art performance (NME) by relatively 10.61\% on average on the RWMB dataset from 9.47 to 8.43, significantly outperforms previous methods by a large margin.

Note that the RWMB is full of real-world motion blur, and thus rather difficult to handle. Most state-of-the-art methods fail to perform well on this dataset for the absence of reliable facial structure information. However, our method manages to remove the motion blur and obtain accurate landmarks.
This experiment demonstrates that our algorithm is robust under real-world motion blur circumstance.

\vspace{-0.2cm}
\subsubsection{Evaluation on 300VW}
\vspace{-0.2cm}
Since most of the video facial landmark detection literature provide testing results on the 300VW dataset, we also evaluate our method on the 300VW dataset to make this paper complete.

\begin{table}
\begin{center}
\begin{tabular}{c|c|c|c|c}
\hline
Method & CateA & CateB & CateC & Challset \\
\hline\hline
REDN~\cite{peng2016recurrent} & - & - & - & 6.25\\
TSTN~\cite{liu2018two} & 5.36 & 4.51 & 12.84 & 5.59\\
Ours & $\pmb {3.56}$ & $\pmb {3.88}$ & $\pmb {5.02}$ & $\pmb{3.96}$\\
\hline
\end{tabular}
\end{center}
\vspace{-0.5cm}
\caption{\label{tab:300vw_origin}Comparison with temporal methods.}
\vspace{-0.5cm}
\end{table}


As is shown in the Figure~\ref{fig:ced_sharp}. There actually exists some motion-blurred frames in category A and C (even though much slight than Blurred-300VW dataset), in which our method largely obtains the state-of-the-art results. Since category B is designed to evaluate the robustness under different illuminations, dark rooms and overexposed shots (which is not the main scenario considered in our work), our method obtains comparable results. Note that comparing to other sophisticated methods, only a vanilla Resnet is used in our model. Generally, this experiment demonstrates that our method does not sacrifice accuracy on regular dataset when obtaining state-of-the-art performance on blurry datasets.

To compare with other methods using temporal information, we report the NME in the Table~\ref{tab:300vw_origin}. Our method works well in the original 300VW dataset and outperforms the existing methods considering temporal information.


\subsection{Ablation Study}

\begin{figure}
\begin{minipage}{0.42\linewidth}
 \centerline{\includegraphics[width=1.3\linewidth]{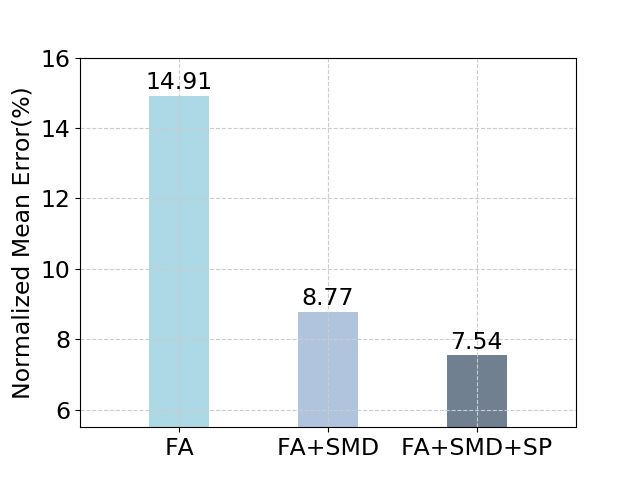}}
 \centerline{(a)}
\end{minipage}
\hfill
\begin{minipage}{0.42\linewidth}
 \centerline{\includegraphics[width=1.3\linewidth]{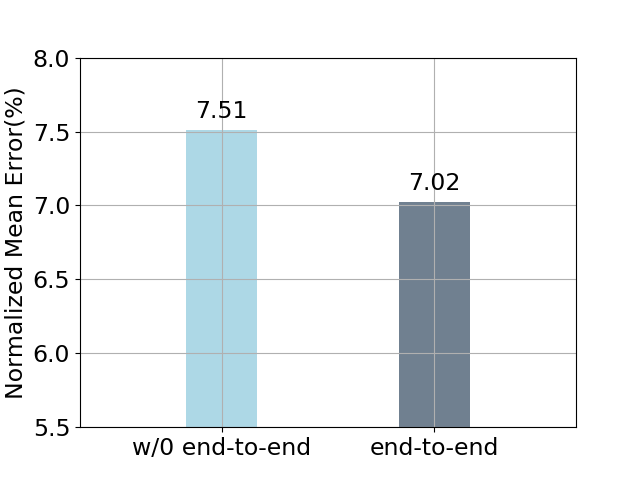}}
 \centerline{(b)}
\end{minipage}
\vspace{-0.2cm}
\caption{\small{(a) Normalized Mean Error to analyze the effect of the pivotal components. (b) Normalized Mean Error to evaluate the effectiveness of alternate fine-tuning strategy.}}
\vspace{-0.2cm}
\label{fig:nme_components_predictor}
\end{figure}

In this section, we investigate the effectiveness of each pivotal component in the proposed framework on the Blurred-300VW Challenging Subset. We still employ the Resnet-18 as the basic facial landmark detector (FA) in our framework. Based on FA, we analyze each proposed component or potential designing, \ie the Structure Predictor (SP), and the Structure-aware Motion Deblurring (SMD), by comparing the NME score. The overall result is presented in Figure~\ref{fig:nme_components_predictor} (a). 
We also analyze the different way to obtain facial edges, \ie by state-of-the-art Structure Detector(SD) or by our Structure Predictor (SP). The comparing results is presented in the Table~\ref{table_detect_vs_predict}.

\begin{figure}[t]
\centering
\includegraphics[width=0.9\linewidth]{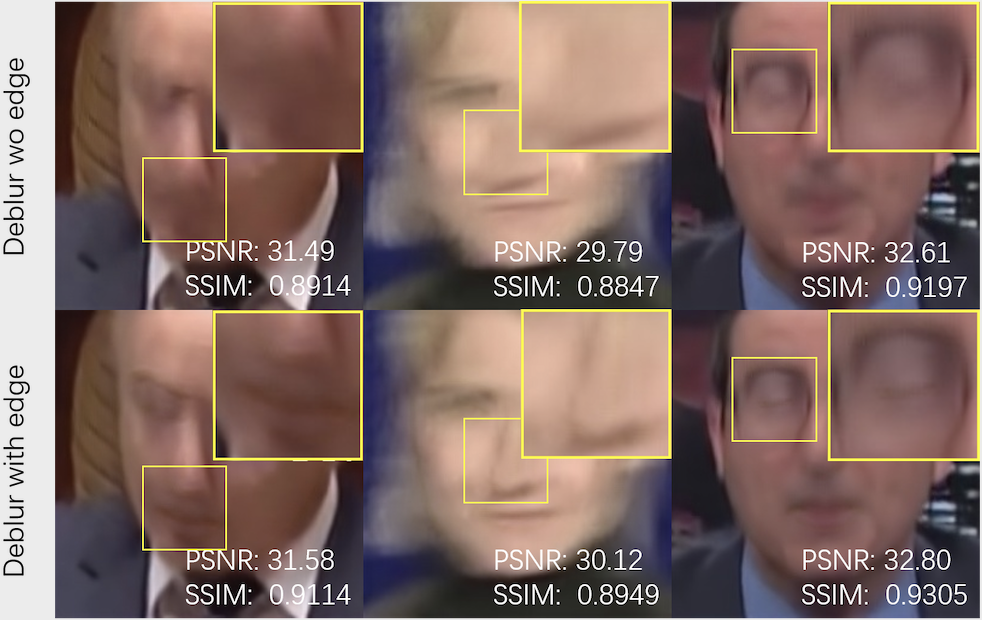}
\vspace{-0.2cm}
\caption{\small{Deblurring w/wo structure information. images in the first row is deblurred by state-of-the-art algorithms, and images in the second row is produced by our algorithm.}}
\vspace{-0.5cm}
\label{fig:psnr}
\end{figure}

\noindent
{\bf Effectiveness of pivotal component.} We first evaluate the effectiveness of the Structure-aware Motion Deblurring (SMD) for facial landmark detection tasks in blurred-videos. The first bar of the Figure~\ref{fig:nme_components_predictor} (a) shows the NME score of the baseline facial landmark detection method fed with blurry images without the assistance of SP or SMD. By adding the state-of-the-art deblurring network~\cite{kim2017online} before facial landmark detector (FA+SMD), we reduce the NME from 14.91\% to 8.77\%, as is shown in the first (FA) and the second bar (FA+SMD) in the Figure~\ref{fig:nme_components_predictor} (a).
Then, we evaluate the effectiveness of Structure Predictor (SP). By leveraging the predicted facial edges with SP, we optimize the NME from 8.77 to 7.54 as is shown in the second (FA+SMD) and the third bar (FA+SMD+SP) of Figure~\ref{fig:nme_components_predictor} (a). As shown in Figure~\ref{fig:psnr}, though motion deblurring is a component rather than the main objective of this work, we still report the comparison of PSNR and SSIM of images deblurred by a state-of-the-art deblurring method~\cite{kim2017online} and our methods with SP. Superior results demonstrate the necessity and the efficiency of our Structure Predictor (SP).

\noindent
{\bf Choice of Structure Predictor.}
One intuitive way to obtain the facial structure is to use a pre-trained state-of-the-art landmark detector. For comparison to our temporal-based structure predictor (SP), we report NME in the Table~\ref{table_detect_vs_predict} of different choices of structure predictor (\ie, FAN\cite{bulat2017far}, ResNet~\cite{he2016identity}, SAN~\cite{dong2018style}, LAB~\cite{wu2018look}). Besides, GT is reported as the upper-bound performance of structure prior usage, in which ground-truth landmarks are used. Our proposed temporal-based structure predictor (SP) obtains the best NME score of 7.54\%.

\noindent
{\bf Replaceable facial landmark detector.}
To verify the feasibility of the replaceable facial landmark detector, we report the NME results in Table~\ref{table_enhance}, which demonstrates that our framework could significantly enhance the performance of single networks by replacing Resnet with them in our framework. It is exciting and encouraging to see that our method enhances most state-of-the-arts significantly.

\noindent
{\bf Alternate fine-tuning strategy.}
After pretraining each component, we end-to-end finetuned them as a whole. As is introduced in Figure~\ref{fig:nme_components_predictor} (b), the baseline (Resnet~\cite{he2016identity}) obtains obvious improvement after the finetuning.

\begin{table}
\begin{center}
\resizebox{0.95\columnwidth}{!}{
\begin{tabular}{c|c|c}
\Xhline{1.2pt}
Row & Method & NME ($\%$) \\
\Xhline{1.2pt}
1 & FA + SMD + SP (FAN~\cite{bulat2017far}) & 8.52 \\
2 & FA + SMD + SP (ResNet~\cite{he2016identity}) & 8.33 \\
3 & FA + SMD + SP (SAN~\cite{dong2018style}) & 8.18 \\
4 & FA + SMD + SP (LAB~\cite{wu2018look}) & 7.92 \\
\Xhline{1.2pt}
5 & \textbf{FA + SMD + SP(Ours)} & \textbf{7.54} \\
6 & FA + SMD + GT & 6.72 \\
\Xhline{1.2pt}
\end{tabular}
}
\end{center}
\vspace{-0.4cm}
\caption{\small{NME of different choices of Structure Predictor as the source of structure prior.}}
\vspace{-0.2cm}
\label{table_detect_vs_predict}
\end{table}

\begin{table}
\begin{center}
\resizebox{0.95\columnwidth}{!}{
\begin{tabular}{c|c|c|c}
\Xhline{1.2pt}
Method & w/o Ours & w/ Ours & Improvement \\
\Xhline{1.2pt}
FAN~\cite{bulat2017far} & 45.42 & 10.95 & 75.89\% $\uparrow$ \\ 
ResNet~\cite{he2016identity} & 14.91 & 7.54 & 49.43\% $\uparrow$ \\
SAN~\cite{dong2018style} & 9.21 & 7.02 & 23.78\% $\uparrow$ \\ 
RFLD~\cite{merget2018robust} & 41.24 & 20.07 & 51.33\% $\uparrow$ \\ 
LAB~\cite{wu2018look} & 8.94 & 4.91 & 45.53\% $\uparrow$ \\ 

\Xhline{1.2pt}
\end{tabular}
}
\end{center}
\vspace{-0.4cm}
\caption{\small{NME(\%) value of state-of-the-art methods, working with or without our framework. Our method enhance these methods efficiently.}}
\vspace{-0.5cm}
\label{table_enhance}
\end{table}

\vspace{-0.3cm}
\section{Discussion and Conclusion}
\vspace{-0.2cm}
In this paper, we introduced an important but omitted issue, the facial landmark detection in motion-blurred videos. Since motion blur is ubiquitous in practical work, we believe further research on this field is of great significance. 
We proposed a novel framework, in which the three components work as an organic whole. The face boundary yielded by the structure predictor assists the deblurring module to relieve the motion blur. And the deblurred face leads to more accurate facial landmarks. There is still a lot of room for improvement for the extreme motion blur of the proposed RWMB dataset. We hope to see further development of this work in the future.



{\small
\bibliographystyle{ieee_fullname}
\bibliography{egbib}
}
\end{document}